\documentclass{article}

\usepackage{microtype}
\usepackage{graphicx}
\usepackage{booktabs} 
\usepackage{booktabs,amsfonts,amsmath,mathtools,framed,pbox,caption,amssymb,subcaption}

\usepackage{amssymb,cite}
\usepackage{epstopdf} 
\usepackage{color}

\usepackage{hyperref}

\usepackage[accepted]{icml2019}

\icmltitlerunning{Modeling and Interpreting Real-world Human Risk Decision Making}

\begin{document}

\twocolumn[
\icmltitle{Modeling and Interpreting Real-world Human Risk Decision Making \\with Inverse Reinforcement Learning}




\begin{icmlauthorlist}
\icmlauthor{Quanying Liu}{caltech,hmri}
\icmlauthor{Haiyan Wu}{cas,caltech}
\icmlauthor{Anqi Liu}{caltech}
\end{icmlauthorlist}

\icmlaffiliation{caltech}{California Institute of Technology, Pasadena, USA}
\icmlaffiliation{cas}{IPCAS, Department of Psychology, University of Chinese Academy of Sciences, Beijing, China}

\icmlaffiliation{hmri}{Huntington Medical Research Institutes, Pasadena, USA}

\icmlcorrespondingauthor{Anqi Liu}{anqiliu@caltech.edu}
\icmlcorrespondingauthor{Haiyan Wu}{wuhy@psych.ac.cn}

\icmlkeywords{Inverse Reinforcement learning, Real-world decision making, Human behavior, Balloon Analogue Risk Taking Task}

\vskip 0.3in
]



\printAffiliationsAndNotice{} 

\begin{abstract}
We model human decision-making behaviors in a risk-taking task using inverse reinforcement learning (IRL) for the purposes of understanding real human decision making under risk. To the best of our knowledge, this is the first work applying IRL to reveal the implicit reward function in human risk-taking decision making and to interpret risk-prone and risk-averse decision-making policies. We hypothesize that the state history (e.g. rewards and decisions in previous trials) are related to the human reward function, which leads to risk-averse and risk-prone decisions.  We design features that reflect these factors in the reward function of IRL and learn the corresponding weight that is interpretable as the importance of features. The results confirm the sub-optimal risk-related decisions of human-driven by the personalized reward function. In particular, the risk-prone person tends to decide based on the current pump number, while the risk-averse person relies on burst information from the previous trial and the average end status. Our results demonstrate that IRL is an effective tool to model human decision-making behavior, as well as to help interpret the human psychological process in risk decision-making.
\end{abstract}

\section{Introduction}
\label{introduction}
Human's real-world decision making under reward and risk has been extensively studied with behavioral tasks and neuroimaging techniques in psychology and neuroscience field. Decision-making tasks are increasingly used to evaluate associations between risk-taking propensity and real-world risky behaviors. Unlike the traditional self-report assessment of risk behavior, the Balloon Analogue Risk Taking Task (BART) is a widely used measure of risk-taking tendency in human.
The BART task was developed by LeJuez et al. \cite{lejuez2002evaluation} as a task consists of balloons that have to be pumped up by individuals. In each trial of the task, the balloon is displayed on the screen with two options for the participant. One option is to stop the pump and secure the amount of money of the current balloon (i.e., cash). An alternative option is to take the risk to pump balloon(i.e., pump), resulting in larger balloon (more rewards) or exploded balloon (losing all accumulated rewards).
The scores in this task have been shown to correlate with self-reported
risk-taking behaviors such as alcohol abuse, drug abuse,
gambling etc \cite{hopko2006construct,lejuez2003evaluation}.

To enhance our understanding of the psychological
process involved in the BART, efforts have been made to decompose the psychological components in BART with computational models. For example, Wallsten et al. proposed the BART-model proposed to quantify the risk-taking, the speed of learning from experience and the behavior consistency~\cite{wallsten2005modeling}. More recently, Van Ravenzwaaij, Dutilh, and Wagenmakers (2011) proposed a 2-parameter (drift rate and non-decision time) simplification diffusion model with empirical data on the BART~\cite{van2012diffusion}. However, these models do not directly capture the representation of the real reward function in risk-taking. 

In this study, we use the inverse reinforcement learning (IRL) to study the real human decision making during the Balloon Analogue Risk Taking Task. IRL aims to find a reward function that explains the observed behavior~\cite{ng2000algorithms,abbeel2004apprenticeship}. Here, we firstly apply IRL on the polling data from all subjects to estimate the weight of each feature as a baseline. We then split the subjects into two groups, risk-prone and risk-averse group, to examine the differences of reward functions of these two groups. Our model can predict the risk-prone person's behavior better. We also find risk-prone person tends to decide based on the current pump number, while the risk-averse person relies on burst information from the previous trial and the average end status.

\section{Related Work}
The risky decisions of human and animal are modulated by a variety of environmental and intrinsic contexts, such as risk propensity and risk perception. One traditional methods considering computational factors for modeling risky behavior of human is under the reinforcement learning (RL) framework \cite{dayan2008reinforcement}. People typically learn the decision-making environments which are characterized by a few key concepts: a state space, a set of actions, and the outcomes from experience.  Applying RL makes it possible to uncover the computational mechanisms and neural substrates for reward prediction error (RPE) and valuation \cite{daw2011model} and show developmental difference \cite{palminteri2016computational} .

Reinforcement learning aims to provide an optimal policy to maximize the rewards, which has been applied to predict the optimal decision policy based on the reward function~\cite{kaelbling1996reinforcement}. However, the reward function in real-world human decision making under risk is usually unknown, for human often tries to avoid the risk states and therefore would not visit all the states. On the other hand, the reward function in our mind might not be the real reward function designed in the task. For instance, the mental status under the risk has to be taken into consideration in human decisions. As a consequence, the real reward function represented in humans might deviate from the true reward function designed in the decision making task. For these reasons, it is still debating in psychology whether humans make rational and optimal decisions under risk. 

Inverse reinforcement learning provides a potential tool to estimate the reward function and learn demonstrator's decision policy~\cite{ng2000algorithms}. While IRL has been implemented to model the route choice ~\cite{ziebart2008maximum}, football players' strategies ~\cite{le2017coordinated}, or robot navigation \cite{kretzschmar2016socially}, to the best of our knowledge, the IRL framework has not been used in psychology and decision-making neuroscience yet.  To recover a reward function of human under risk and learning policies from risk-prone and risk-averse people, IRL predicts actions in states which even has not been observed/occurred.

\section{Task and Behavioral Data}
\label{experiment}

\subsection{Balloon Analogue Risk Task}
We used the Balloon Analogue Risk Task (BART), to assess the real-world human decision-making behavior through the conceptual frame of balancing the potential for reward versus risk~\cite{lejuez2002evaluation}. The instructions were the same as the ones described by LeJuez et al. (2002). Participants were asked to earn as many points as possible. 

\begin{figure}[ht]
\begin{center}
\centerline{\includegraphics[width=0.85\columnwidth]{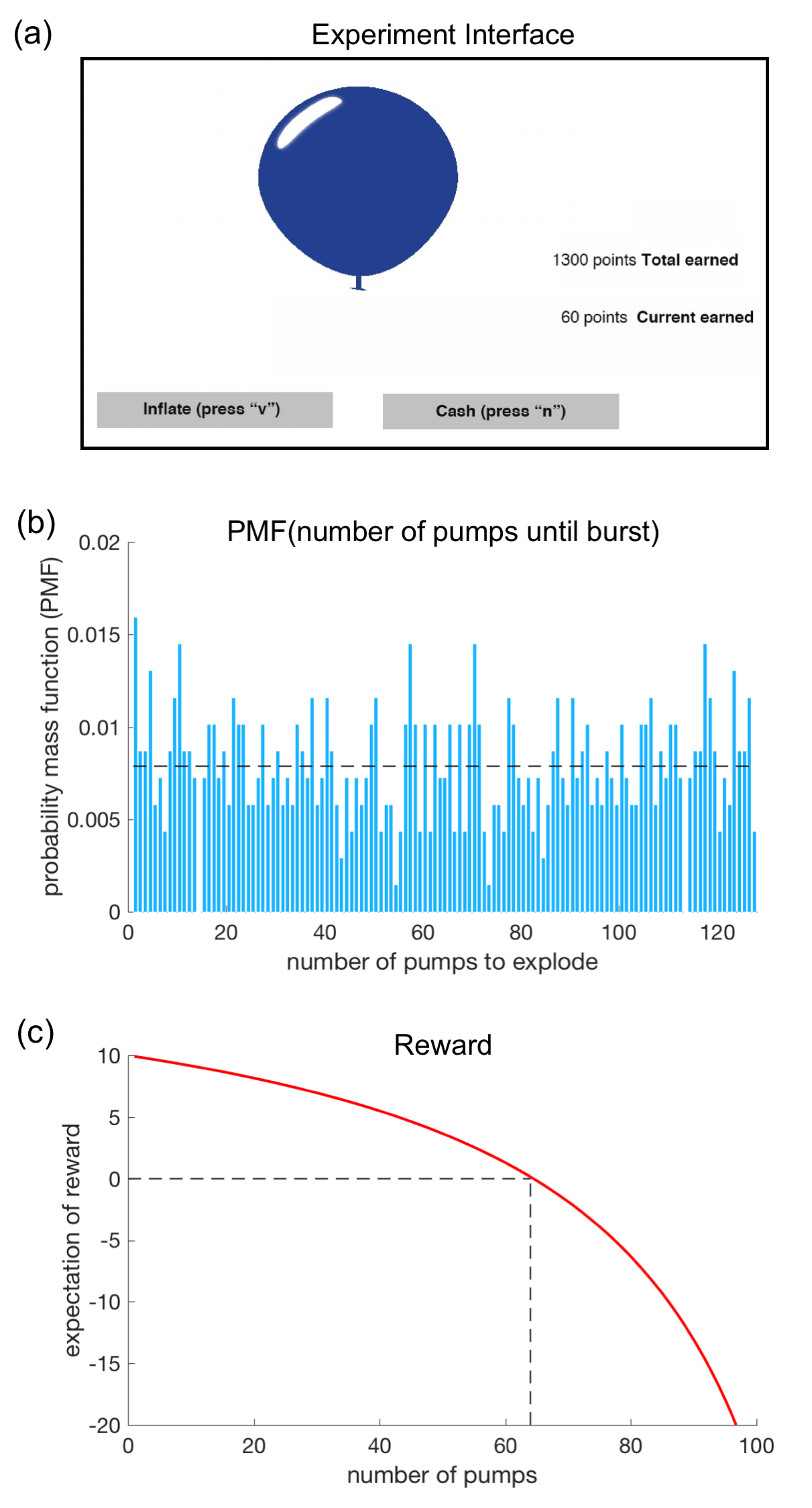}}
\caption{Experiment settings. (a) The experimental interface; (b) the probability mass function for the number of pumps until balloon explosion; (c) the true expectation of reward.}
\label{fig:exp}
\end{center}
\vskip -0.2in
\end{figure}

The experimental interface is in (Figure \ref{fig:exp}a). Subjects can choose to pump the balloon (press ``v") or stop pump (press ``n"). As a result of choosing to pump, the balloon may burst or may grow in size. If the balloon burst, the participant loses all the accumulated points in the current trial, otherwise the participant earns 10 more points and confronts with a next-round choice: cash out or pump. A new trial begins when the participant stops pump (cashes out) or the balloon bursts. The task consists of 1 practice trial and 30 formal task trials.

In each trial, the breakpoints are uniformly distributed from 1 to 128 (Figure \ref{fig:exp}b). It means the probability that a balloon would explode on the first pump is 1/128. If the balloon did not blow on the first pump, the probability of a balloon burst on the second pump is 1/127, and so on. Based on this setting, the true reward function at i\textsuperscript{th} pump is
\begin{equation}
\label{eq:true_reward_function}
\begin{aligned}
R(i)=
&\begin{dcases}  -10(i-1), & \text{if exploded }  \\
    +10, & \text{otherwise } .
 \end{dcases}
\end{aligned}
\end{equation}
Let $X_i=1$ denote the event that the balloon burst at i\textsuperscript{th} trial. We have $P(X_i=1|X_{i-1}=0)=\frac{1}{129-i}$, and $P(X_i=0|X_{i-1}=0)=\frac{128-i}{129-i}$. 
Therefore, the expected reward at i\textsuperscript{th} pump is
\begin{equation}
\label{eq:expected_reward_function}
\begin{aligned}
E(i) = \frac{1290-20i}{129-i}
\end{aligned}
\end{equation}
The expected reward starts from 10 at first trial, and decreases with the increasing number of trials. The best strategy for this task is to cash out at 64\textsuperscript{th} pump, when the reward reduces to 0 (Figure \ref{fig:exp}c).

\subsection{Subjects and Human Behaviors}
We collect experimental data with extensive coverage and report simple statistics. There are 23 university students (15 females with $M_{age} = 23.2$, $SD_{age} = 3.22$) from Beijing,
aged 18 to 25, participating in the experiment and the monetary reward varies upon the task performance.

\begin{figure}[ht]

\begin{center}
\centerline{\includegraphics[width=0.9\columnwidth]{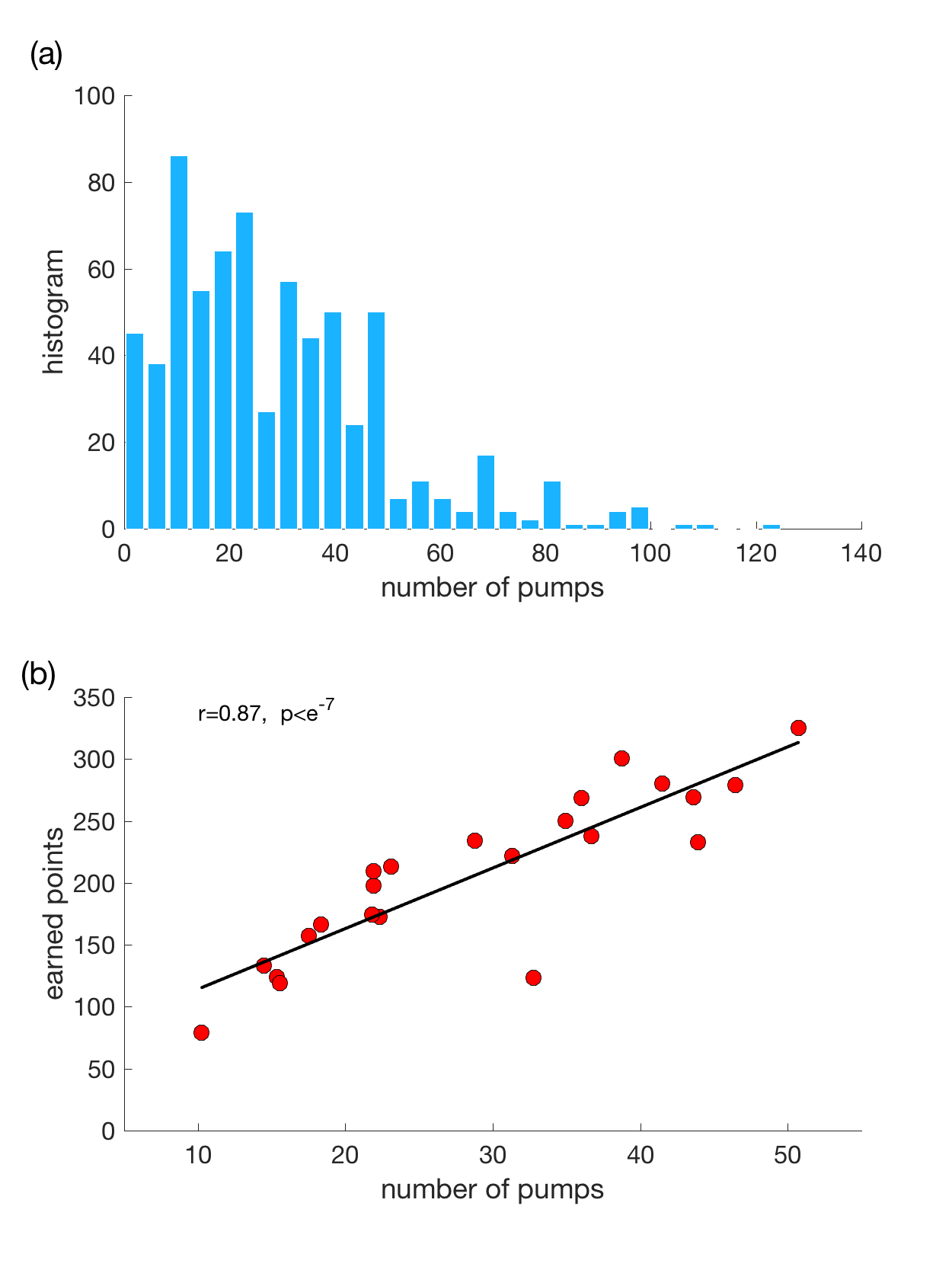}}
\caption{Behavioral data. (a) the histogram of number of pumps; (b) reward as a function of number of pumps.}
\label{fig:behavior}
\end{center}
\vskip -0.2in
\end{figure}

The behavioral results indicate a mean of pump number as 29.04 , which is much smaller than a rational player would pump. The participants stopped pumping and cashed out before the balloon exploded in 74.14\% of trials. The mean reward points in each test earned by the participants were 205.92. The histogram of the number of pumps for all the trials are in Figure \ref{fig:behavior}a. It seems that the participants did not learn the true reward function during the task, for they stopped to pump before the expected reward decrease to 0. As a consequence, the more pumps, the more rewards. The final rewards are significantly correlated with the number of pumps (Figure \ref{fig:behavior}b). 

\section{Algorithm}
We model the human behavior in this risk-taking task using Markov Decision Process. We argue that the Markovian property of transition and action holds in this particular task with certain assumptions when setting up state and action space. Specifically, we set states as two dimensional vector $[i,s]$, where $i$ represent the $i\textit{th}$ pump and $s$ represents the current status of the balloon as {0,1}, meaning burst or not. Therefore, we can assume the next state, $[i+1, s]$, meaning the $(i+1)\textit{th}$ pump and the corresponding status of the balloon only depend on the previous status and current action. The transition relation shown in Eq.~\ref{eq:true_reward_function} also reflects this relation. The policy $P(a_t|s_1, s_2, ...,s_t)$ can be also regarded as equivalent to $P(a_t|s_t)$ since the human subject always know how many inflation of the balloon has been executed. Our action space is binary with 0 indicating ``stop" while 1 indicating ``pump".

In BART, human subjects resemble agents in a reinforcement learning task. The goal of the human subjects is to collect the largest reward from a single trial without knowing the probability of balloon burst. We can regard this process as human learning of an unknown reward function. Because of the irrational nature of human and their different risk-prone levels, the true reward function (burst probability) is never learned. Instead, the human decision making in this process provides data that demonstrate possible underlying influence of historical trials to current ones. In other words, the data is not independent with each other between different trials. However, we construct the state features in a way that reflects information from historical trials and we make the assumption that state features is independent among trials. Table~\ref{tab:features} shows our state features. Note that by ``stop" we mean ending the trial without balloon burst while by ``end" we mean ending the trial with whatever status.

%

\begin{table}
\begin{tabular}{|p{0.5cm}|p{1cm}|p{5.5cm}|}
\hline
 No.  & Value & Meaning \\ \hline
 
 1 & Integer &\scriptsize{Number of times being in this state} \\ \hline
 2 & Binary &\scriptsize{Whether this state was burst in the previous trial}\\ \hline
 3 & Binary &\scriptsize{Whether this state was stopped in the previous trial}\\ \hline
 4 & Binary &\scriptsize{Whether state was burst in the 2nd previous trial}\\ \hline
 5 & Binary &\scriptsize{Whether this state was stopped in the 2nd previous trial}\\ \hline
 6 & Binary & \scriptsize{Whether state was burst in the 3rd previous trial}\\ \hline
 7 & Binary &
\scriptsize{Whether this state was stopped in the 3rd previous trial}\\ \hline
 8 & Binary &
\scriptsize{Whether is the average burst status}\\ \hline
 9 & Binary &
\scriptsize{Whether is the average stop status}\\ \hline
 10 & Binary &
\scriptsize{Whether is the average end status} \\ \hline
 11 & Integer & \scriptsize{Number of steps in the current trial}\\
\hline

\end{tabular}
\caption{States Features with Value and Meaning.}
\label{tab:features}
\end{table}

\subsection{Maximum Entropy IRL}

Traditional Markov decision process (MDP) assumes that we know the reward function of agents. However, in the real world, humans usually make decisions without knowing the reward function. In imitation learning, we observe demonstrations of expert trajectories. IRL reduces learning to a problem of recovering utility function that makes the behavior induced by a near-optimal policy closely mimic the demonstrated behavior ~\cite{ziebart2008maximum, abbeel2004apprenticeship}. More specifically, IRL assumes reward function, $R$, has a linear relation with specific features that are associated with states representation, and the reward value of a trajectory, $\boldsymbol{f}(\zeta)$, is simply the sum of state rewards $\boldsymbol{f}(s_t)$:
\begin{align}
    R(\boldsymbol{f}(\zeta)) 
    = \theta^T \boldsymbol{f}(\zeta) 
    = \sum_{s_t\in \zeta} \theta^T \boldsymbol{f}(s_t), \label{eq:reward}
\end{align}
where $\theta$ is the reward weights. A deep neutral network can be also used for representing the relation between features and reward function. 

In the most prevalent IRL model, which is the MaxEntIRL, the model finds the policy that maximizes the entropy of trajectories:
    $\theta = arg\max_{\theta} L(\theta) = arg\max_{\theta} \sum_{\text{examples}} \log P(\tilde{\zeta}|\theta, T)$,
where a convenient form of policy distribution is derived in inference, possessing the form:
\begin{align}
    P(a_t|s_t) \propto \exp(\theta^T\boldsymbol{f}(s_t)),\label{eq:policy}
\end{align}
where $\theta$ is the learning parameters (reward weights) and $\boldsymbol{f}(s_t)$ is the state features. We follow the Algorithm 1 in ~\cite{ziebart2008maximum} for inference of the state frequencies. The learning objective is the approximate likelihood of the samples collected from human experiments. Therefore, the maximum entropy IRL algorithm finds reward weights that makes human behavior appear near optimal under the learned reward function. It resolves the ambiguity from the real-world actions and resulting in a single stochastic policy.  The gradient is the difference between expected features frequencies and empirical ones. We can use gradient descent to optimize for parameter $\theta$.



\section{Experiments}
\label{results}
We conduct two sets of experiments to demonstrate how IRL can recover reward function underlying human decision-making behavior and help neuroscientist and psychologist analyze how historical observations and risk-taking habits can affect future decision making. We first use all the merged data together and regard everyone as the same type of risk-taking pattern and learn the reward function. We then compare with the result we obtain in group-specific experiments where we split subjects into two groups: risk-prone group and risk-averse group, based on the median pump numbers in pooling data. Table \ref{tab:test_groupstat} shows the different behavior patterns of these two groups of people. Risk-prone group tends to have longer trial length, meaning they pump balloon much more times and have longer total time. But they have a shorter reaction time per pump, meaning they make a decision quicker. However, the risk-averse group tends to have shorter trial length, shorter total time, but longer reaction time for making decisions. 
For each group, we use half of the data for training and the other half for testing.
\begin{table}
\begin{tabular}{|p{1.7cm}|p{1.7cm}|p{1.7cm}|p{1.75cm}|}
\hline
 & All subjects & Risk-Prone & Risk-Averse \\ 
 \hline
 \scriptsize{Mean (\# pumps)} & 29.04 & 38.78 & 18.42 \\
 \hline 
 \scriptsize{Mean (RT/trial)} & 13.77 & 16.20 & 11.11\\
 \hline
 \scriptsize{Mean (RT/pump)} & 0.53 & 0.46 & 0.60 \\
 \hline
 \end{tabular}
 \caption{Number of bumps and reaction time for different groups. }
 \label{tab:test_groupstat}
\vskip -0.2in
 \end{table}

\subsection{Reward Function Analysis}
\begin{figure}[ht]
\vskip 0.2in
\begin{center}
\centerline{\includegraphics[width=0.95\columnwidth]{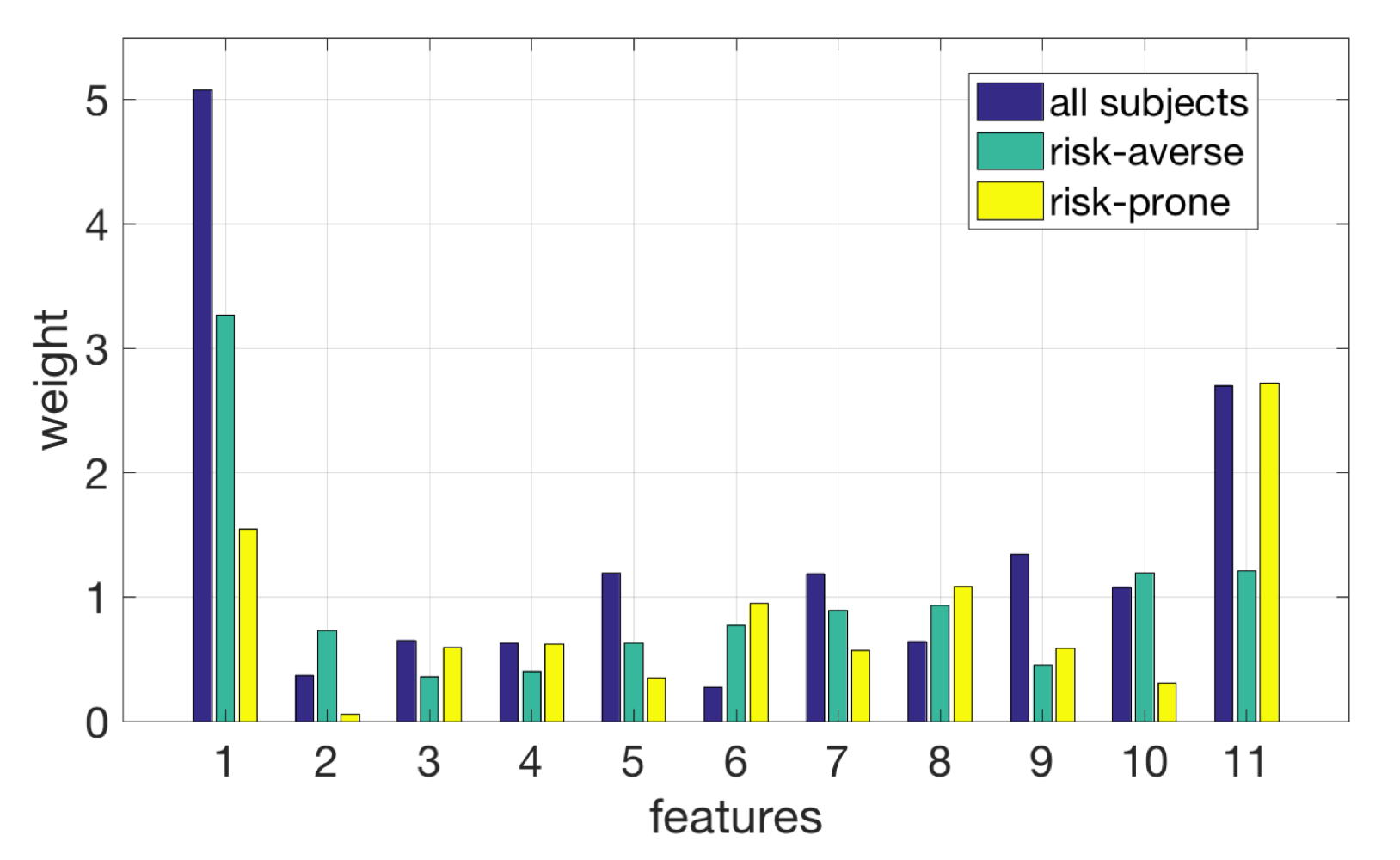}}
\caption{Weights of features from IRL model. The meaning for the 11 features are listed in Table \ref{tab:features}}
\label{fig:weights}
\end{center}
\vskip -0.2in
\end{figure}

The weights of features for all subjects, risk-averse subjects and risk-prone subjects, are in Figure \ref{fig:weights}. 
Overall, people show higher weight of ``Number of times being in this state" (feature 1), indicating a general effect of past individual experience on the current risky decision making \cite{bechara2000emotion,hertwig2009description}. 
Interestingly, risk-prone subjects tend to weigh more on ``Number of steps in the current trail" (feature 11), with less weight on ``Number of times being in this state" (feature 1), ``Whether burst in this state in the previous trial" (feature 2) and ``Whether is the average end status" (feature 10), than risk-averse subjects. Our findings suggest that the history of "Number of times being in this state" serves as a general cue of contextual analysis when making a risky decision in outcome anticipation. Irrationally, the current number of pumps is more influential in risk-prone subjects than risk-averse individuals. 
Risk-averse subjects rely more on the previously learned information (from both the previous trial and the average end status).
Thus, our results support the idea that people are learning to be risk-averse, with consideration of normal human as adaptive decision makers \cite{march1996learning,denrell2007adaptive}.

\subsection{Predictive Performance Analysis}
We test the learned model on testing data and demonstrate the average likelihood of the test trajectories in Table \ref{tab:test_lld}. The likelihood of data sequence for a trial $\eta$ is as following using the same model:
\begin{align}
P(\eta) = P(s_0)P(a_0|s_0)P(s_1|a_0, s_0)P(a_1|s_1)...
\end{align}

\begin{table}
\begin{tabular}{|p{1.5cm}|p{1.7cm}|p{1.7cm}|p{1.75cm}|}
\hline
 & All subjects & Risk-Prone & Risk-Averse \\ 
 \hline
 Test LLD & -1.36&-0.89 & -1.4\\
 \hline
 \end{tabular}
 \caption{Test Log Likelihood for Different Groups. The higher log likelihood, the better prediction of IRL is. }
 \label{tab:test_lld}
 \end{table}
The predictive results (Table \ref{tab:test_lld}) indicate that the decisions from the risk-prone group are easier to predict than the risk-averse group. Convergently, risk-prone individuals also show less hesitation (shorter reaction time per pump) when deciding whether to pump the balloon (Table \ref{tab:test_groupstat}), compared to the risk-averse people.

\section{Discussions}
\label{discussion}
This work aims to model human reward function in the BART under the framework of IRL. With the construction of IRL models for all subjects, risk-prone subjects and risk-averse subjects respectively, we reveal the weights for different features for each group.

\subsection{Psychological Perspective}

Our results (Figure \ref{fig:behavior}) reveal a general risk averse behavioral pattern as most people did not reach the optimal pump point (64 pumps in our task). Such finding is consistent with existing results showing risk aversion in normal population \cite{loewenstein2001risk,fullenkamp2003assessing,holt2002risk}. It implies that human subjects did not search the entire state space and therefore did not learn the true reward function (Figure \ref{experiment}c). 

Other alternative models have been used to learn and predict the human decision makings~\cite{angela2009sequential,ryali2018demystifying,guo2018so}, such as dynamic belief model which requires knowledge or assumption about the models, or reinforcement learning which assume humans know the reward function. 
The key concept of IRL here is that humans make decisions to optimize an unknown reward function which is linear in the features. Our results quantified the weight of features in the reward function represented in human's decision, and show that risk-prone person tends to decide based on the current bump number, while the risk-averse person rely on burst information
from previous trial and the average end status (Figure \ref{fig:weights}). Moreover, we found the IRL can predict the risk-prone subject's decisions better than the risk-averse subject's decisions (Table \ref{tab:test_lld}), suggesting the reward functions in risk-prone subjects are highly representative. This might result in high  impulsivity and less variety in their decisions under risk, which has been shown in the literature ~\cite{kashdan2009social,adriani2009increased}. 

\subsection{Machine Learning Perspective}
Inverse Reinforcement Learning is a prevalent tool for imitation learning and learning from demonstration~\cite{ng2000algorithms}, which has been applied successfully to robotics, self-driving car, and assistive technology~\cite{zhifei2012survey}. This work reveals the potential advantage for using IRL to study human decision making in psychological experiment with limited samples and large variability. Comparing with simpler models like behavior cloning~\cite{ross2011reduction}, IRL can recover reward function even when only limited states are visited in data. But it also induces a slightly larger computational cost. The comparison between different machine learning models is an interesting research topic that we leave to future work. Importantly, this work is also an exploration towards interpretable machine learning when applying machine learning to fields that require high-level model interpretability. When we analyze predictive performance of the model, we do not seek the best model for highest accuracy but the best one providing the most reasonable explanation of data. Finally,
the limited data of psychological experiments can heavily constrain the model choice for machine learning. This drives new developments of domain adaption and active learning techniques in machine learning community.

\begin{figure}[ht]
\begin{center}
\centerline{\includegraphics[width=1.1\columnwidth]{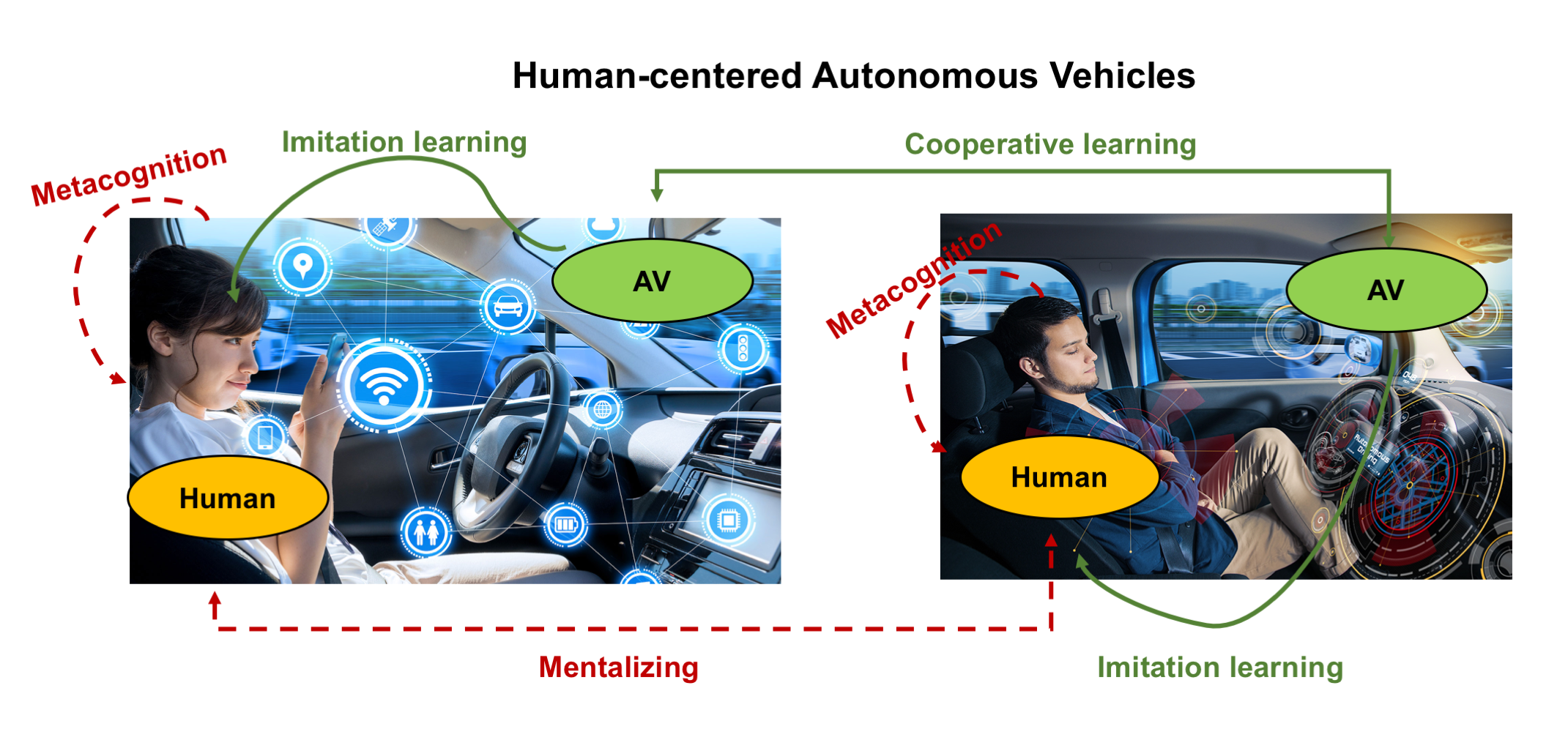}}
\caption{Learning human's cost function in autonomous vehicles}
\label{fig:AV}
\end{center}
\vskip -0.2in
\end{figure}

\subsection{Future Directions}
In the future, we will improve the model by integrating the individual personalities from questionnaires (e.g. sensation seeking, and impulsivity)\cite{donohew2000sensation,lauriola2014individual,bornovalova2009risk} and other contextual factors (such as risky cue learning) \cite{cohen2009neuroelectric} into our feature space.
Moreover, we only tested 30 trials for each subject, which is not enough to train a model for each subject, and personalize the reward function. We will run more trials per subject, train IRL model and other computational models in the future. It will allow us to examine the cross-subject variation in model parameters. The implications of the IRL model in risk decisions can also benefit both understanding risk-taking behavior at an individual level or interactive mind level (e.g. metacognition or mentalizing process for the mental state inference within/between individuals), which can guide the design of human-centered autonomous vehicles (Figure \ref{fig:AV}).

\section*{Acknowledgements}
QL gets supports from James Boswell Fellowship and FWO postdoctoral fellowship (12P6719N LV). HW gets supports from the National Natural Science Foundation of China (grant number: U1736125).



\bibliography{example_paper}
\bibliographystyle{icml2019}

\end{document}